\def\year{2021}\relax
\documentclass[letterpaper]{article} % DO NOT CHANGE THIS
\usepackage{aaai21}  % DO NOT CHANGE THIS
\usepackage{times}  % DO NOT CHANGE THIS
\usepackage{helvet} % DO NOT CHANGE THIS
\usepackage{courier}  % DO NOT CHANGE THIS
\usepackage[hyphens]{url}  % DO NOT CHANGE THIS
\usepackage{graphicx} % DO NOT CHANGE THIS
\usepackage{natbib}  % DO NOT CHANGE THIS AND DO NOT ADD ANY OPTIONS TO IT
\usepackage{caption} % DO NOT CHANGE THIS AND DO NOT ADD ANY OPTIONS TO IT

\usepackage{pdfpages} % DO NOT CHANGE THIS
\usepackage{animate}

\newcommand{\animtool}[6]{
 \begin{figure}[h]
 \centering
 \begin{frame}{}
   \animategraphics[loop,controls,autoplay,width=#1\linewidth]{#2}{#3}{#4}{#5}
 \end{frame}
 \caption{[Animation] #6}
 \end{figure}
}

\usepackage{amsmath}
\usepackage[switch]{lineno} 

\newcommand{\nullitem}{}
\newcommand{\npcite}[1]{\citeauthor{#1} \citeyear{#1}}
\newcommand{\npnguyen}{\citeauthor{nguyen:ijcnn19} \citeyear{nguyen:ijcnn19}}

\newcommand{\mytitle}{Emergence of Different Modes of Tool Use in a Reaching and Dragging Task}

\newcommand{\myauthorblock}{
	\author{
	  Khuong Nguyen and Yoonsuck Choe\\
	}
	\affiliations{
	        Department of Computer Science and Engineering,\\
		Texas A\&M University, College Station, TX 77843-3112\\
		khuong.nguyen1289@gmail.com, choe@tamu.edu
	
	}
} % end myauthorblock

\title{\mytitle}

\myauthorblock

\newcommand{\getfig}[2]{\includegraphics[width=#1\columnwidth]{#2}}
\newcommand{\threeplot}{0.3}

\newcommand{\fnhplot}{0.245}
\newcommand{\sixplot}{0.4}

\newcommand{\tabsummary}{
\begin{table}[h]
\centerline{
\begin{tabular}{|l||l|l|l|}  			\hline
 & T-shaped & L-shaped & I-shaped \\ 		\hline \hline
Success rate & {62\%} & 50\% & 45\% \\ 		\hline
Reward & {943.2} & 806.2 & 644.7 \\		\hline
Episode Length & {280.92} & 302.51 & 374.56 \\ 	\hline
\end{tabular}
}
\caption{Results by Tool Type (mean over 100 episodes)}
\label{t:summary}
\end{table}
}
\newcommand{\tablshape}{
\begin{table}[h]
\centerline{
\begin{tabular}{|l||l|l|}  					\hline   
  & I-like use & T-like use \\  	\hline \hline
Upper Arena & 3.33\% & 52.5\%  \\ 	\hline
Lower Arena & 36.67\% & 7.5\%  \\ 	\hline \hline
Overall & 40\% & 60\% \\ 		\hline
\end{tabular}
}
\caption{L-shaped Tool Use Modes when Tip of L is Facing Away from Object}
\label{t:lshape}
\end{table}
}
\newcommand{\tabishape}{
\begin{table}[h]
\centerline{
\begin{tabular}{|l||l|l|l|}  					\hline    
     & Wide sweep& Hitting & Other \\ \hline \hline 
Upper Arena & 55.56\% & 2.22\% & 15.56\% \\ 	\hline
Lower Arena & 15.56\% & 8.89\% & 2.22\% \\ \hline \hline
Overall & 71.11\% & 11.11\% & 17.78\% \\ 		\hline
\end{tabular}
}
\caption{I-shaped Tool Use Modes}
\label{t:ishape}
\end{table}
}
\newcommand{\tabthrow}{
\begin{table}[h]
\centerline{
\begin{tabular}{|l||l|l|l|}  					\hline    
     & Correcting & Throwing & Others \\ \hline \hline 
T-shaped & 9.68\% & 0.00\% & 90.32\% \\ \hline
L-shaped & 6.00\% & 2.00\% & 92.00\% \\ \hline
I-shaped & 6.67\% & 11.11\% & 82.22\% \\ \hline
\end{tabular}
}
\caption{Throwing and Correcting Behavior by Tool Type}
\label{t:throw}
\end{table}
}
\newcommand{\figarchi}{
\begin{figure}[h]
\centerline{\getfig{1.0}{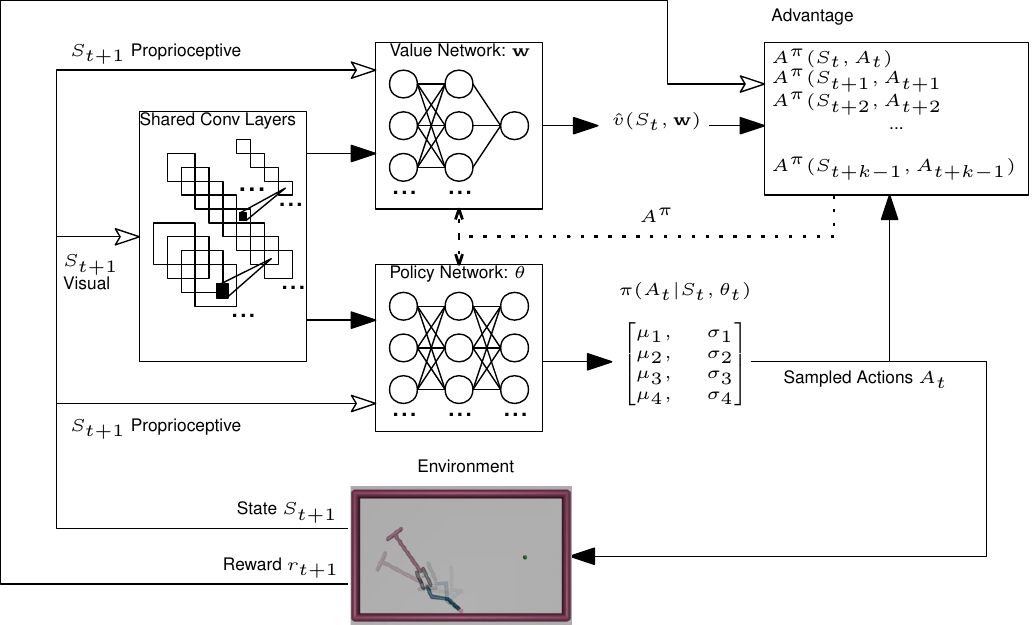}}
\caption{Task Environment and Model Architecture.}
\label{f:archi}
\end{figure}
}
\newcommand{\figtrain}{
\begin{figure*}[t]
\centerline{
\footnotesize
\begin{tabular}{c@{}c@{}c@{}c@{}c}
\getfig{\sixplot}{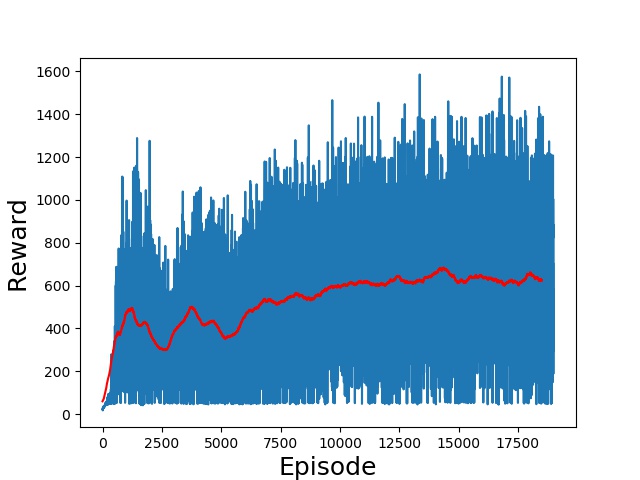} &
\getfig{\sixplot}{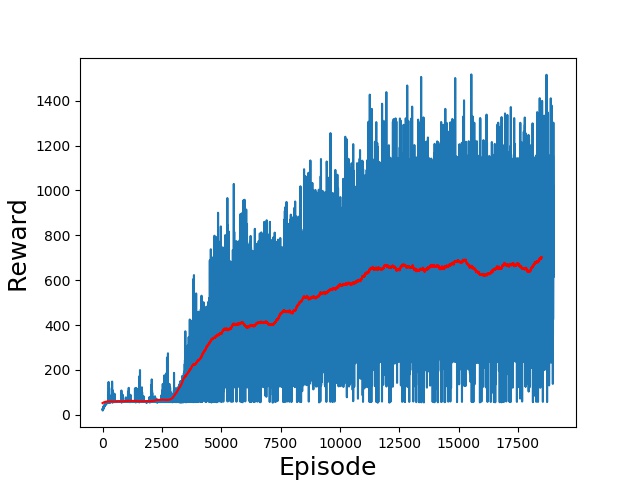} &
\getfig{\sixplot}{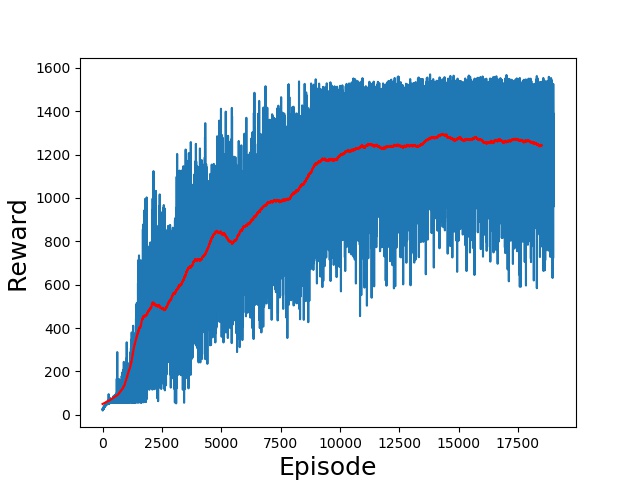} &
\getfig{\sixplot}{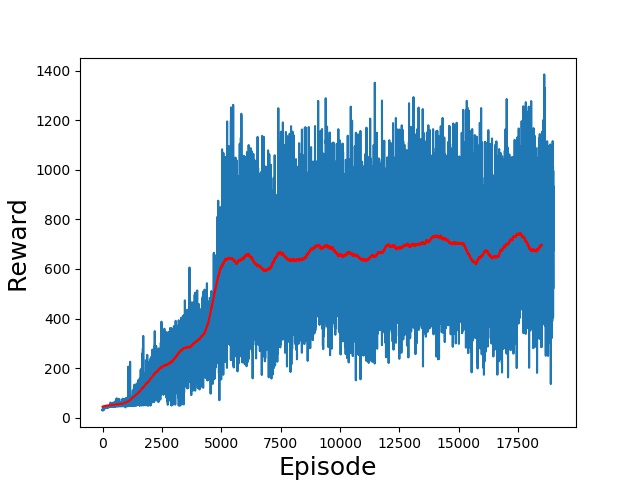} &
\getfig{\sixplot}{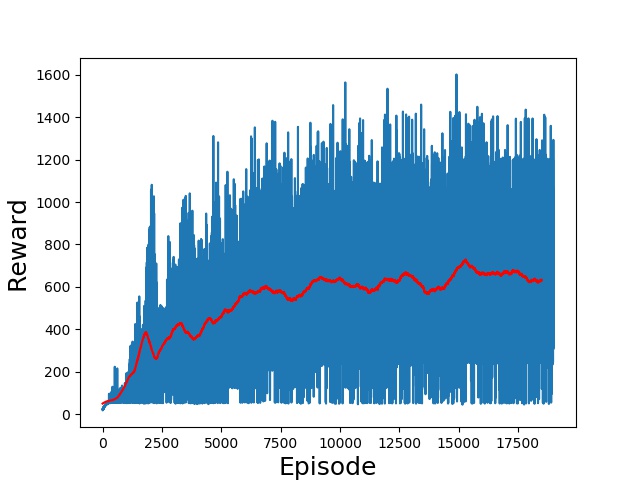} \\
\getfig{\sixplot}{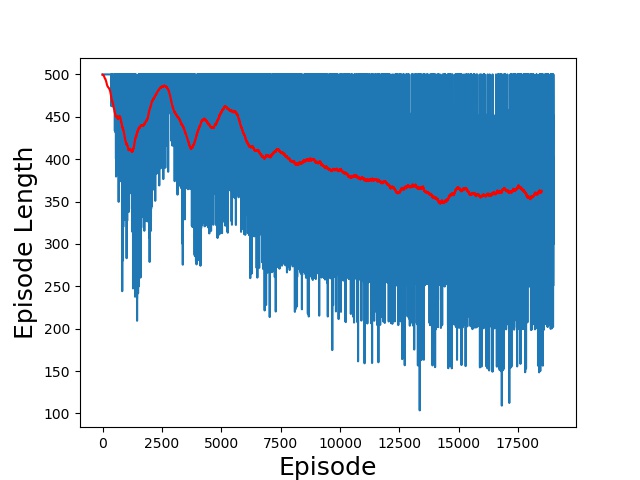} &
\getfig{\sixplot}{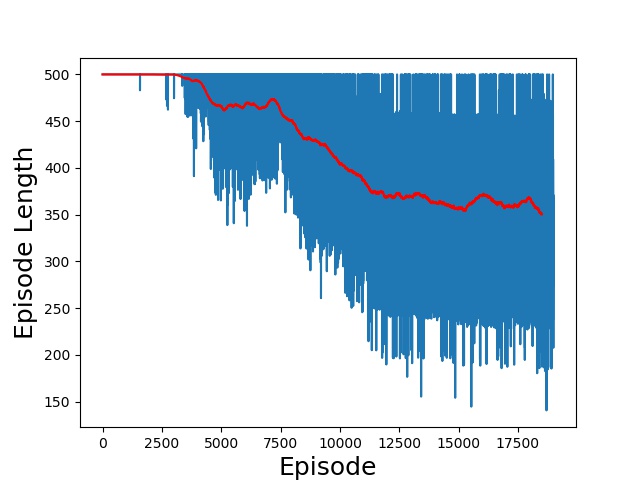} &
\getfig{\sixplot}{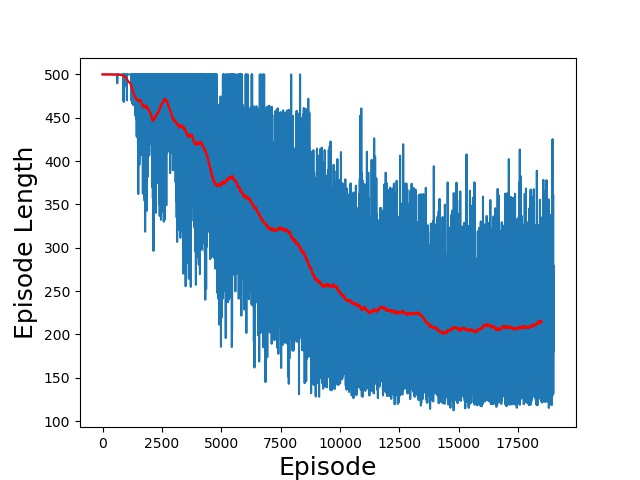} &
\getfig{\sixplot}{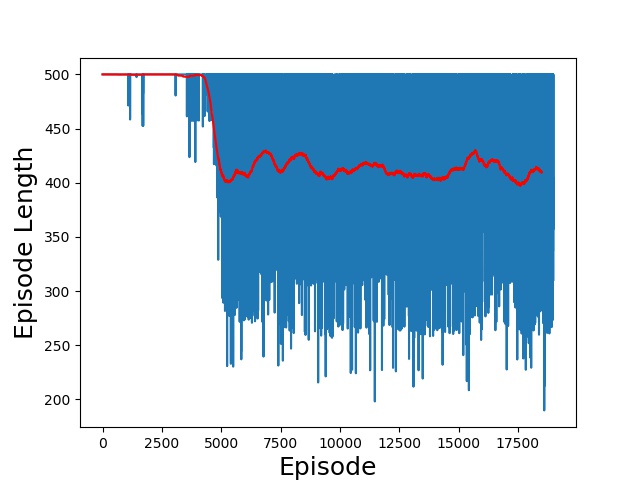} &
\getfig{\sixplot}{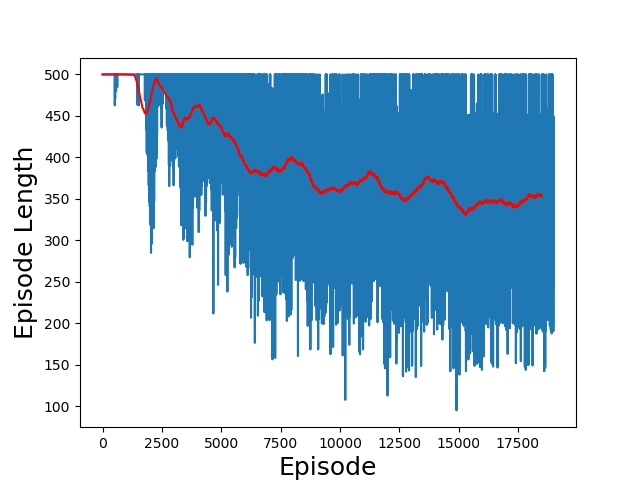} \\
Trial 1 & Trial 2 & Trial 3 & Trial 4 & Trial 5 \\ % & Average \\
\end{tabular}
}
\caption{Training Results: (Top) Reward. (Bottom) Episode Length. Running average and actual reward/episode length.}
\label{fig:perf_five_runs_part1}
\label{f:train}
\end{figure*}
}
\newcommand{\figtdrag}{
\begin{figure}[h]
\centerline{
\footnotesize
\begin{tabular}{ccc}
\getfig{\threeplot}{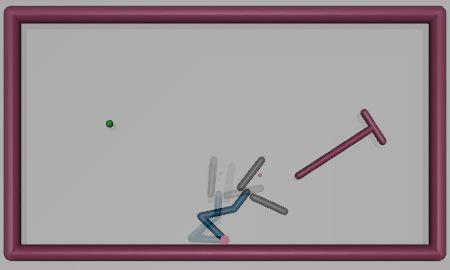} & 
\getfig{\threeplot}{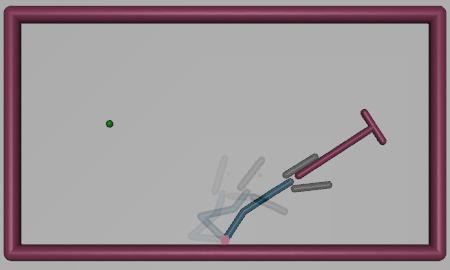} & 
\getfig{\threeplot}{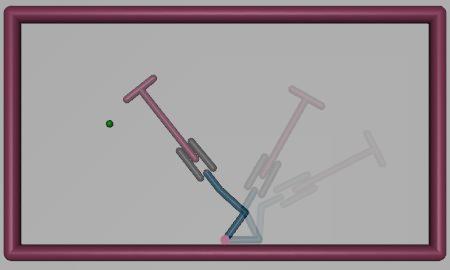} \\ 
t =  8 & t =  11 & t =  25 \\
\getfig{\threeplot}{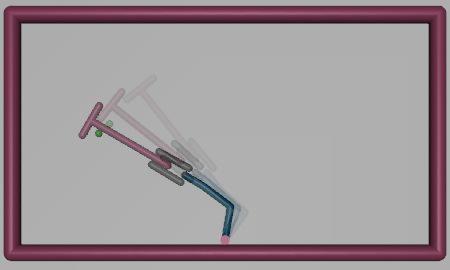} & 
\getfig{\threeplot}{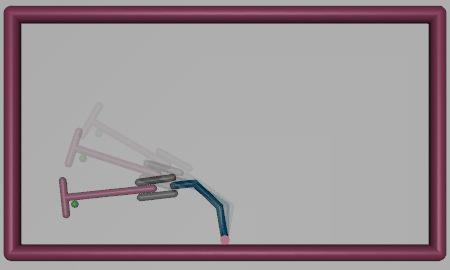} & 
\getfig{\threeplot}{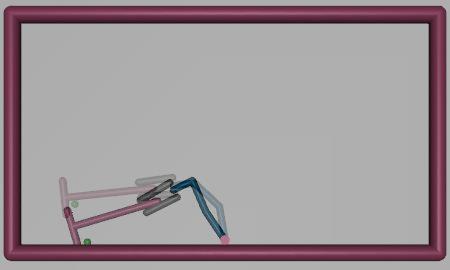} \\ 
t =  27 & t =  29 & t =  30\\
\end{tabular}
}
\caption{T-shaped Tool: Typical Behavior (The semi-transparent overlays show immediate preceding frames.)}
\label{f:tdrag}
\end{figure}
}
\newcommand{\figlmode}{
\begin{figure}[h]
\centerline{
\footnotesize
\begin{tabular}{ccc}
\getfig{\threeplot}{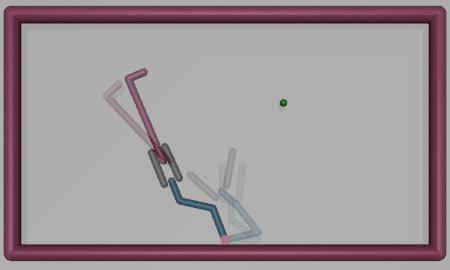} & 
\getfig{\threeplot}{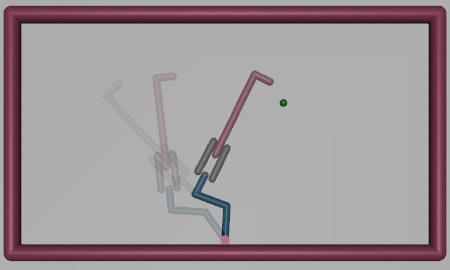} & 
\getfig{\threeplot}{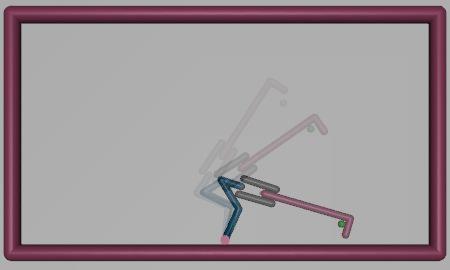} \\
 t =  10 & t =  23 & t =  29\\
\getfig{\threeplot}{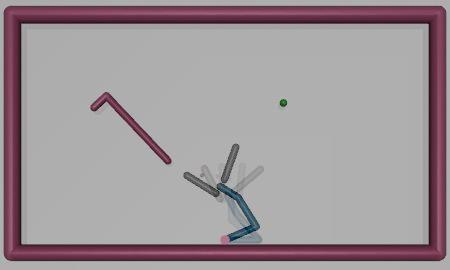} & 
\getfig{\threeplot}{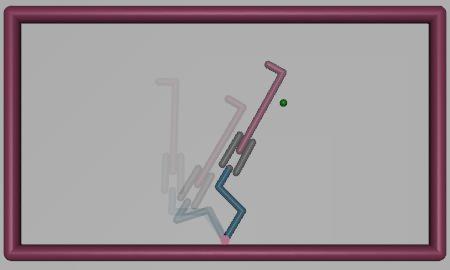} & 
\getfig{\threeplot}{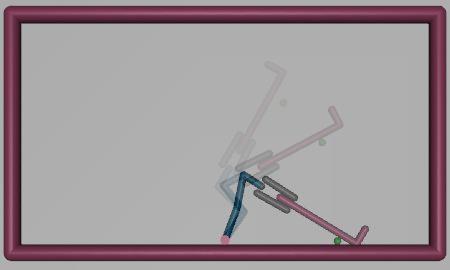} \\ 
 t =  5 & t =  26 & t =  30\\
\end{tabular}
}
\caption{L-shaped Tool: Tool Tip Direction Relative to Object. (Top) Tip of L points toward the object (T-like). (Bottom) Tip of L points away from the object (I-like).}
\label{f:lmode}
\end{figure}
}
\newcommand{\figlcorrect}{
\begin{figure}[h]
\centerline{
\footnotesize
\begin{tabular}{ccc}
\getfig{\threeplot}{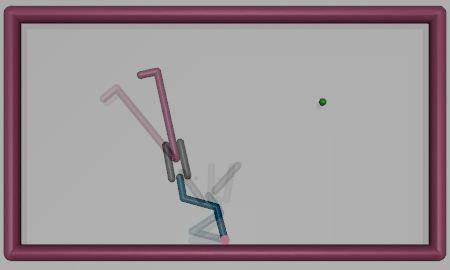} & 
\getfig{\threeplot}{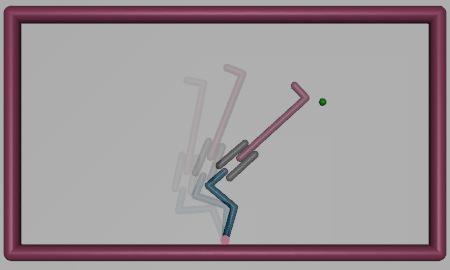} & 
\getfig{\threeplot}{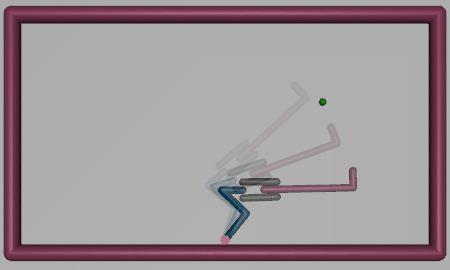} \\
t =  7 & t =  13 & t =  15\\
\getfig{\threeplot}{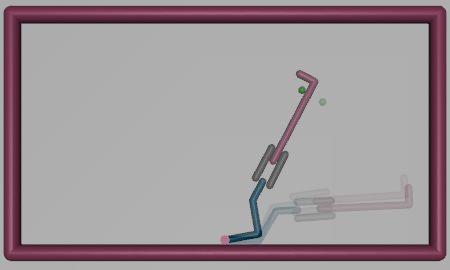} & 
\getfig{\threeplot}{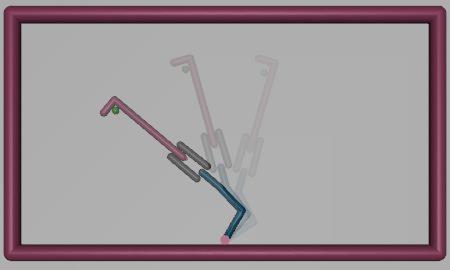} & 
\getfig{\threeplot}{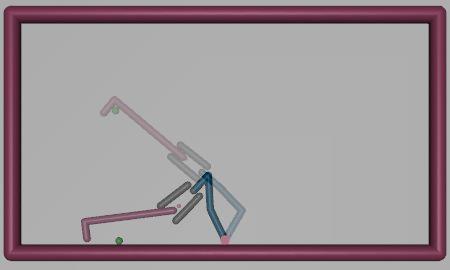} \\
t =  26 & t =  29 & t =  30\\
\end{tabular}
}
\caption{L-shaped Tool: Maneuvering Tip Around Object.}
\label{f:lcorrect}
\end{figure}
}
\newcommand{\figiwide}{
\begin{figure}[h]
\centerline{
\footnotesize
\begin{tabular}{ccc}
\getfig{\threeplot}{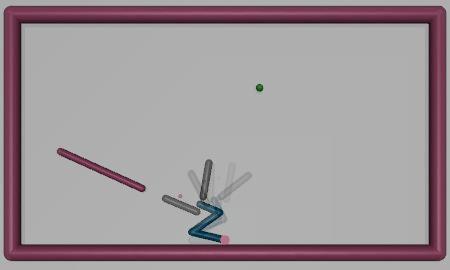} & 
\getfig{\threeplot}{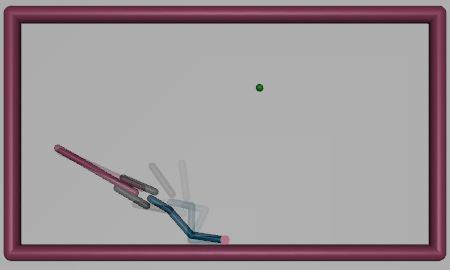} & 
\getfig{\threeplot}{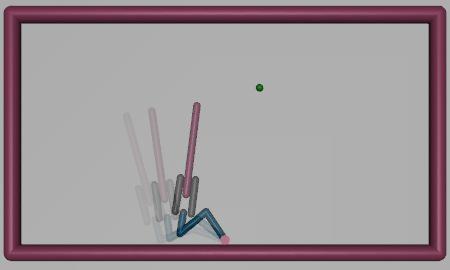} \\
t =  3 & t =  6 & t =  10\\
\getfig{\threeplot}{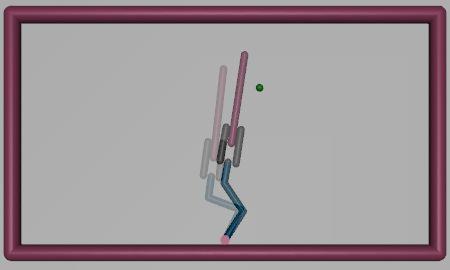} & 
\getfig{\threeplot}{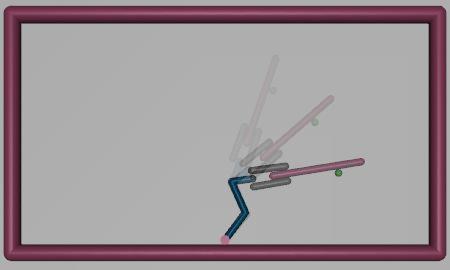} & 
\getfig{\threeplot}{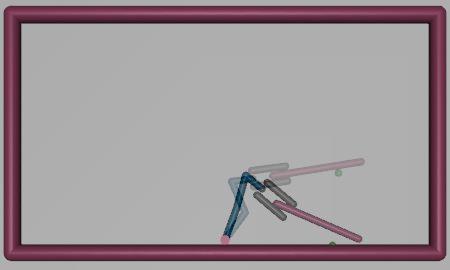} \\
t =  12 & t =  15 & t =  16 \\
\end{tabular}
}
\caption{I-shaped Tool: Wide Sweep Behavior}
\label{f:iwide}
\end{figure}
}
\newcommand{\figihit}{
\begin{figure}[h]
\centerline{
\footnotesize
\begin{tabular}{ccc}
\getfig{\threeplot}{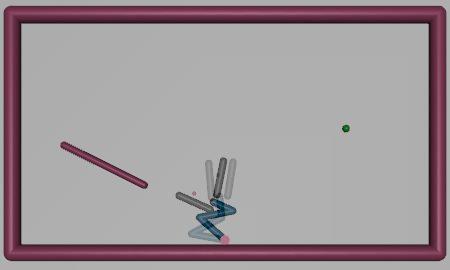} & 
\getfig{\threeplot}{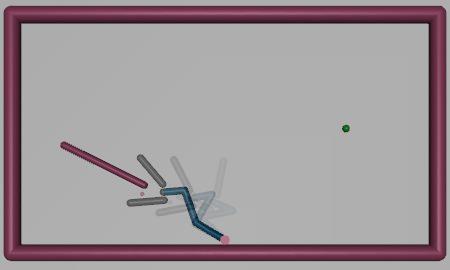} & 
\getfig{\threeplot}{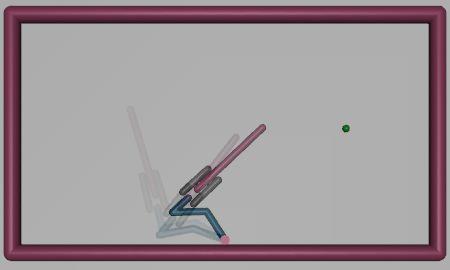} \\
t =  3  & t =  6  & t =  7 \\
\getfig{\threeplot}{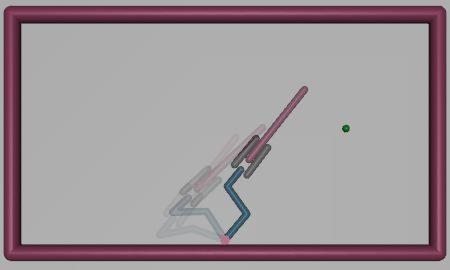} & 
\getfig{\threeplot}{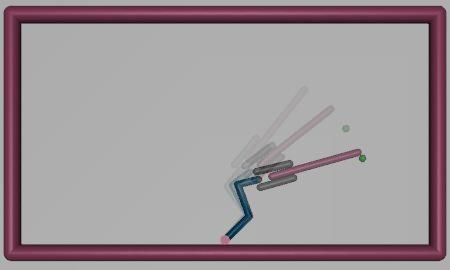} & 
\getfig{\threeplot}{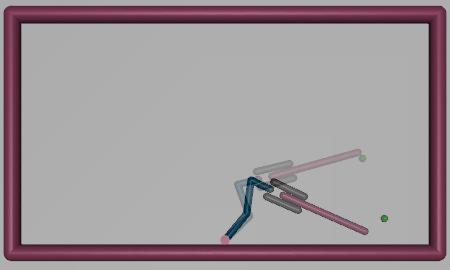} \\
t =  12 & t =  14 & t =  15\\
\end{tabular}
}
\caption{I-shaped Tool: Hitting Behavior}
\label{f:ihit}
\end{figure}
}
\newcommand{\figithrow}{
\begin{figure}[h]
\centerline{
\footnotesize
\begin{tabular}{ccc}
\getfig{\threeplot}{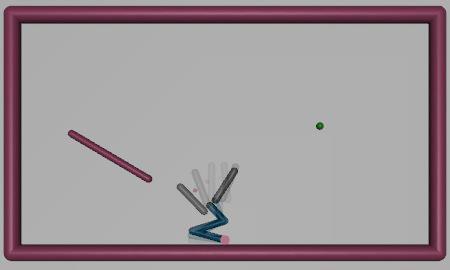} &
\getfig{\threeplot}{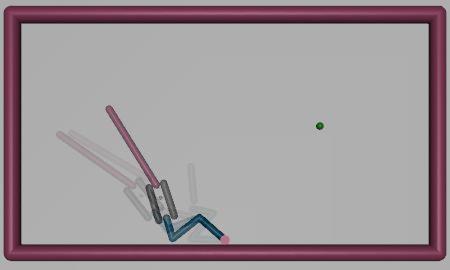} &
\getfig{\threeplot}{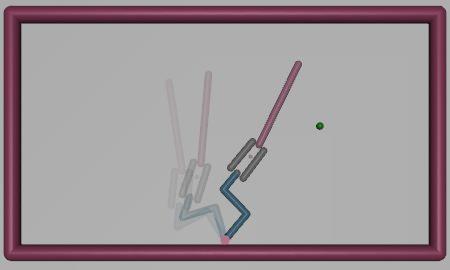} \\
t =  3& t =  9 & t =  13\\
\getfig{\threeplot}{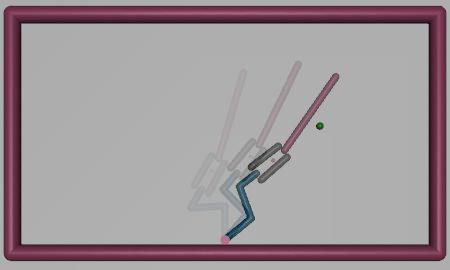} &
\getfig{\threeplot}{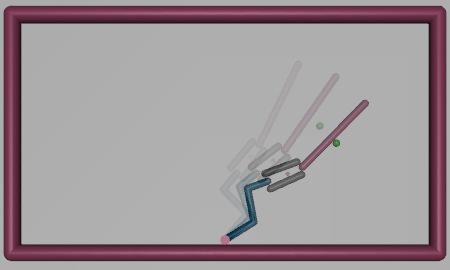} &
\getfig{\threeplot}{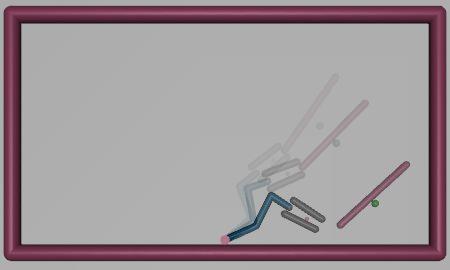} \\
t =  14 & t =  15& t =  16\\
\end{tabular}
}
\caption{I-shaped Tool: Throwing Behavior}
\label{f:ithrow}
\end{figure}
}
\newcommand{\figlcorrecttwo}{
\begin{figure}[h]
\centerline{
\footnotesize
\begin{tabular}{ccc}
\getfig{\threeplot}{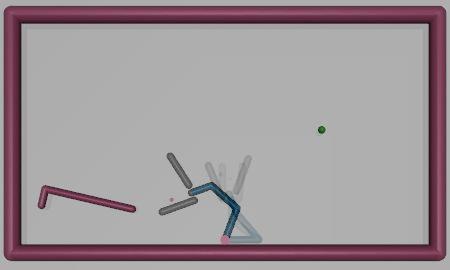} & 
\getfig{\threeplot}{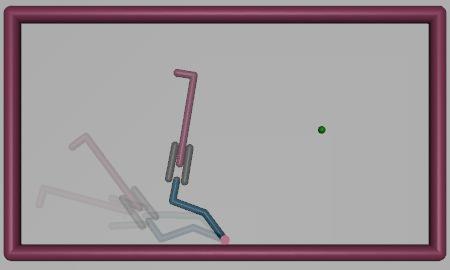} & 
\getfig{\threeplot}{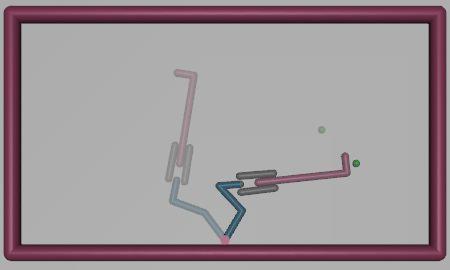} \\ 
t =  4 & t =  9  & t =  10 \\
\getfig{\threeplot}{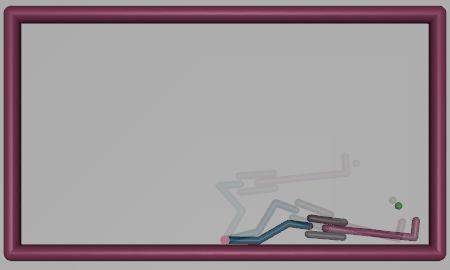} & 
\getfig{\threeplot}{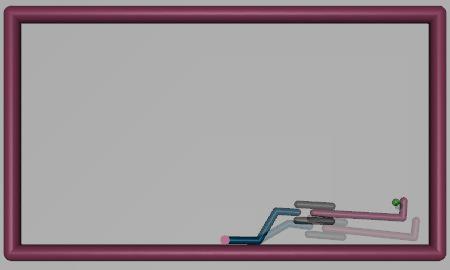} & 
\getfig{\threeplot}{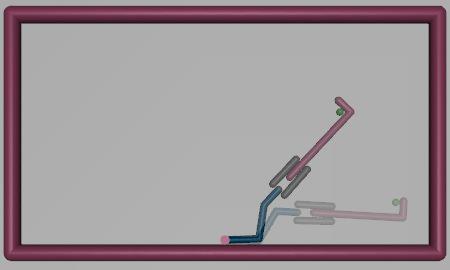} \\ 
t =  12 & t =  13 & t =  14\\
\getfig{\threeplot}{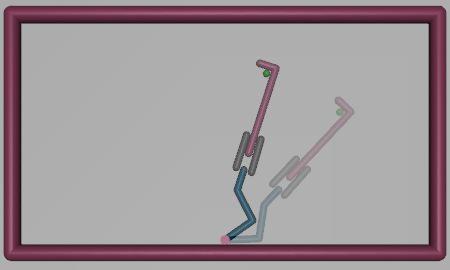} & 
\getfig{\threeplot}{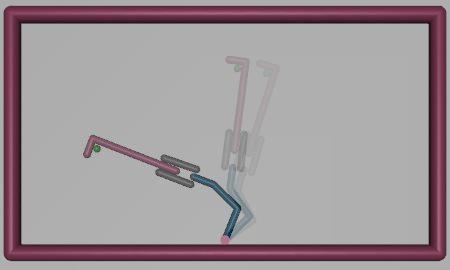} & 
\getfig{\threeplot}{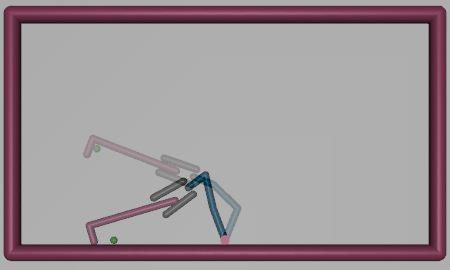} \\ 
t =  15  & t =  17 & t =  18\\
\end{tabular}
}
\caption{L-shaped Tool: Correcting Behavior.}
\label{f:lcorrecttwo}
\end{figure}
}
\newcommand{\figfail}{
\begin{figure}[h]
\centerline{
\footnotesize
\begin{tabular}{@{}c@{}c@{}c@{}c@{}}
\getfig{\fnhplot}{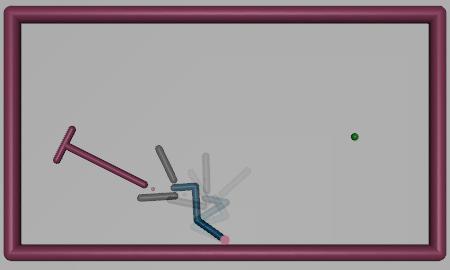} & 
\getfig{\fnhplot}{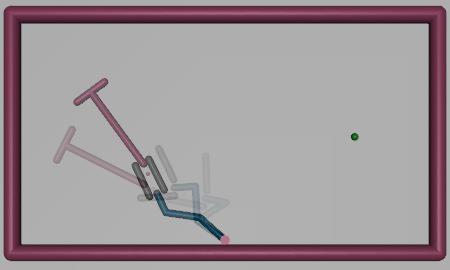} & 
\getfig{\fnhplot}{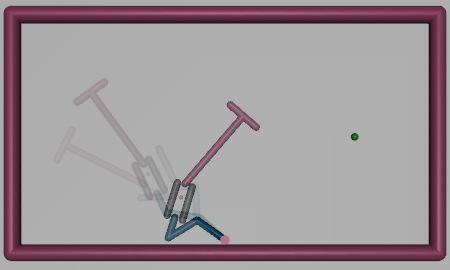} & 
\getfig{\fnhplot}{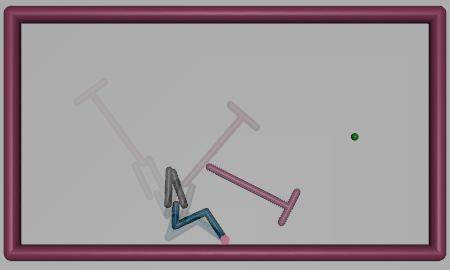} \\
t =  4 & t =  5& t =  6& t =  7\\
\getfig{\fnhplot}{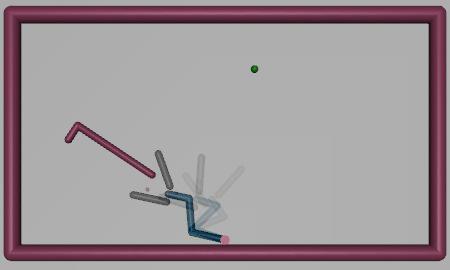} & 
\getfig{\fnhplot}{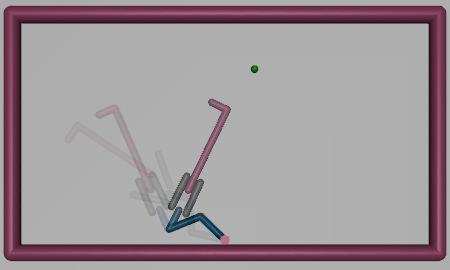} & 
\getfig{\fnhplot}{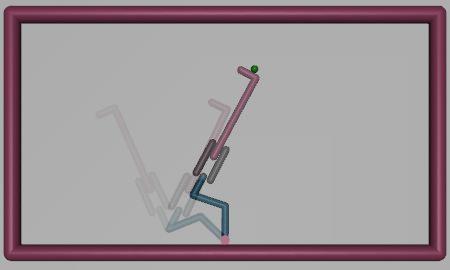} & 
\getfig{\fnhplot}{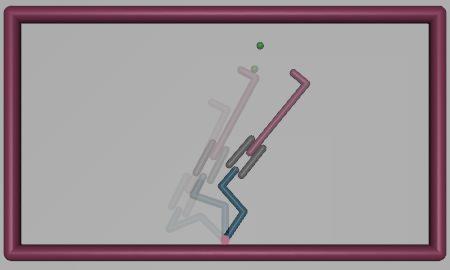} \\
t =  4 & t =  6& t =  7& t =  8\\
\getfig{\fnhplot}{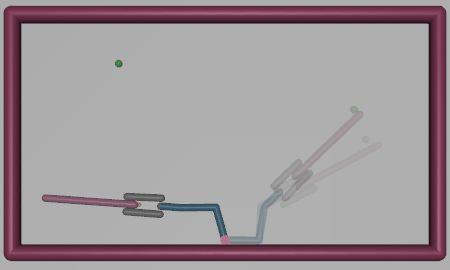} & 
\getfig{\fnhplot}{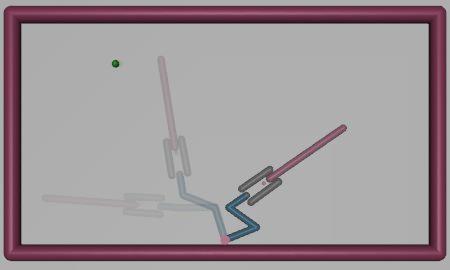} & 
\getfig{\fnhplot}{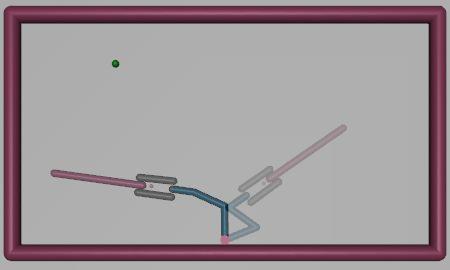} & 
\getfig{\fnhplot}{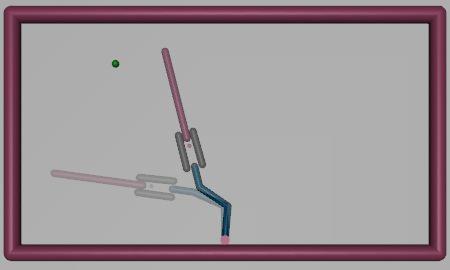} \\ 
t =  4 & t =  6& t =  7& t =  8\\
\end{tabular}
}
\caption{Failure Modes: (Top) Lose grasp of tool. (Middle) Nudge object away.
(Bottom) Flailing.
\label{f:fail}
}
\end{figure}
}
\begin{document}

%\linenumbers 

\maketitle

%%%%%%%%%%%%%%%%%%%%%%%%%%%%%%%%%%%%%%%%%%%%%%%%%%%%%%%%%%%%%%%%%%%%%%%%%%%%
% main part :  aaaaaa
%%%%%%%%%%%%%%%%%%%%%%%%%%%%%%%%%%%%%%%%%%%%%%%%%%%%%%%%%%%%%%%%%%%%%%%%%%%%
\begin{abstract}

Tool use is an important milestone in the evolution of intelligence.  In this paper, we investigate different modes of tool use that emerge in a reaching and dragging task. In this task, a jointed arm with a gripper must grab a tool (T, I, or L-shaped) and drag an object down to the target location (the bottom of the arena). The simulated environment had real physics such as gravity and friction. We trained a deep-reinforcement learning based controller (with raw visual and proprioceptive input) with minimal reward shaping information to tackle this task. We observed the emergence of a  wide range of unexpected behaviors, not directly encoded in the motor primitives or reward functions. Examples include hitting the object to the target location, correcting error of initial contact, throwing the tool toward the object, as well as normal expected behavior such as wide sweep. Also, we further analyzed these behaviors based on the type of tool and the initial position of the target object. Our results show a rich repertoire of behaviors, beyond the basic built-in mechanisms of the deep reinforcement learning method we used.

\end{abstract}

%%%%%%%%%%%%%%%%%%%%%%%%%%%%%%%%%%%%%%%%%%%%%%%%%%%%%%%%%%%%%%%%%%%%%%%%%%%%
% Intro
%%%%%%%%%%%%%%%%%%%%%%%%%%%%%%%%%%%%%%%%%%%%%%%%%%%%%%%%%%%%%%%%%%%%%%%%%%%%
\section{Introduction}

The ability to solve a task using different types of tools indicates high cognitive function, including the ability to reason and perceive what is best to use for a job. In non-human species, Corvids such as the New Caledonian crows have been observed to not only be able to use different types of tools \cite{holzhaider2008wild} but also to make new tools from existing materials to gain access to food \cite{mehlhorn2010tool,von2018compound}. Primates such as chimpanzees and Capuchin monkeys have also been seen to exhibit such behavior \cite{sirianni2015choose,luncz2016wild}.

In artificial intelligence research, there is an increasing interest in tool use. Work by \citet{nishide2011tool} focused on the tool-body assimilation aspect where they used self-organizing maps to extract the information from visual input to identify different type of tools, followed by recurrent neural networks for learning the dynamics, and a hierarchical neural network for tool-body assimilation (this work served as the main motivation for our work reported here). \citet{saponaro2017learning} used a visual and motor mechanism to represent different hand postures to generalize tool affordance learning (for affordance, see \npcite{gibson1977theory}). Recent work of \citet{xie2019improvisation} demonstrated a robot that can learn from demonstration to use novel objects as an improvisational tool to complete a simple sweeping task. \citet{wicaksono2016relational} used inductive logic programming to model the learning but the study is limited in that each tool is learned separately and the trained robot can only handle one type of tool for each trained model. \citet{baker2019emergent} showed that with standard deep reinforcement learning, together with multi-agent competition, can exhibit rich collaborative tool use behavior, even when the game rule was very simple. However, in their work, the tools were boxes, ramps, and walls, and the only mode of interaction with the tool was pushing and pulling these objects.

In our paper, we investigate tool use behavior using standard deep RL algorithms, as in \citet{baker2019emergent}. One key difference with \citeauthor{baker2019emergent} is that our agent is a jointed limb with a gripper which has to grasp a tool and wield it to reach and drag an object, so that the agent has to be more deliberate than simple pushing/pulling. Our main focus is on how different affordances offered by different types of tools (T-, L-, and I-shaped tools) and the environment affect the behavior of the deep RL agent (in our previous work, we considered T-shaped tool only, and did not include extensive behavioral analysis: \citeauthor{nguyen:ijcnn19} \citeyear{nguyen:ijcnn19}). We show that, with standard deep RL and simple rewards, the agent can exhibit novel behaviors such as hitting, throwing, and exaggerated movement to compensate for the shortcoming of specific tools. Our quantitative analysis indicates that these emergent behaviors are largely necessitated by the different affordances of the tools and the task situation.

\section{Methods}

\subsection{Task Environment}

%This improved environment is built upon the original tool use environment with an increased level of complexity. Besides the stability improvement of the simulation, two new important features, including the use of different types of tool and random tool position are added. 

The environment introduces (similar to our prior work \npnguyen) a robot with a three-joint arm with one gripper (which has its own joint), three types of tools (T-, L-, and I-shaped tools [only T-shaped in \npnguyen]), and a small target object, all enclosed in a rectangular arena (figure \ref{f:archi}). The task of the agent is to try to grasp the tool and use it to drag the object down to the target area, which is the bottom of the arena. The target is initially placed beyond the reach of the gripper, so tool use is necessary. 

Each state $S$ (the input) at time $t$ provided to the agent by the environment consists of 4 consecutive snapshots of the environment: 4 images and 4 proprioceptive and/or kinesthetic feedbacks (joint position and angular velocity). The visual inputs are acquired using the image frames of the scenes, and the proprioceptive/kinesthetic feedbacks are computed from the information provided by OpenAI's MuJoCo-py API. Pixel inputs are normalized by dividing by 255, and proprioceptive feedbacks are normalized by using running estimates of their means and standard deviations. 

The movements were in a continuous action space. The actions include four continuous values where three of them are the torques that apply to the three joints on the agent's limb. The fourth action value is applied to the gripper joint and considered as a discrete value where 1.0 is applied to close the gripper if the action input for the joint is greater than 0, otherwise -1.0 is applied to open the gripper. Since the actions are in continuous space, their values are produced by sampling multidimensional normal distributions (4 in this case) with $\mu_i$ and $\sigma_i$ being the mean and standard deviation of the $i$-th normal distribution, generated by the reinforcement learning algorithm (the policy network).

\figarchi 

Three types of tools, including the T-shaped, I-shaped, and L-shaped tool were used, following \citet{nishide2011tool}. During testing, the type of tool is arbitrarily chosen and placed randomly in the environment, within reach of the agent in each running episode. The tools appear with equal probability to prevent any bias or preference for a specific tool. The target object position is generated randomly so that it is within a certain distance from the randomly located tool. These features are expected to increase the complexity of the environment and therefore increase the difficulty level to solve the task. The environment was implemented using the MuJoCo physics simulator (friction, force, etc. modeled: \citeauthor{todorov2012mujoco} \citeyear{todorov2012mujoco}) and OpenAI Gym \cite{brockman2016openai}.  

\subsection{Actor-Critic Based Controller}

The description below closely follows our prior work \cite{nguyen:ijcnn19}. For our deep RL controller, we used a modified version of the Actor-Critic using Kronecker-Factored Trust-Region (ACKTR) algorithm. ACKTR was introduced by \cite{wu2017scalable} where the method showed higher performance than other state-of-the-art reinforcement learning approaches such as A2C \cite{mnih2016asynchronous}, PPO \cite{schulman2017proximal}, and TRPO \cite{schulman2015trust}. The modified version of ACKTR we used turns the original ACKTR into a synchronous ACKTR \cite{pytorchrl}. This method gives us the benefit of A2C, where it creates multiple versions of the agent to interact with multiple versions of the environment to learn more efficiently. At the same time, it still preserves all the advantages that the original ACKTR has. 

%Our Actor-Critic architecture is organized into two parts. The first part is shared between the actor network and the critic network which includes 3 convolutional layers (32@8x8, 64@4x4, and 32@3x3) and one linear layer (512 units) with a Leaky ReLU activation function comes with each layer. The output of the linear layer is concatenated with the proprioceptive input to feed into the second component. The second component consists of the individual parts of the Actor-network with three linear layers (64 neurons, 64 neurons, and 4 output neurons) and the individual parts of the critic network with another three linear layers (64 neurons, 64 neurons, and 1 output neuron) with a tanh activation function comes with each layer on the actor branch and a Leaky ReLU activation function on the critic branch. Two normalization units are also used to normalize the data before they are feed into the actor and critic network.

The Actor-Critic architecture we used is shown in figure \ref{f:archi}. The Actor and Critic share the same block that contains 3 Conv+Pooling layers (55$\times$48 RGB image input [4 frames], 32 channels with 8$\times$8 receptive field [RF], 64 channels with 4$\times$4 RF, 32 channels with 3$\times$3 RF) followed by a 512-neuron fully connected (FC) layer, each with Leaky ReLU activation. The Actor model (policy network) consists of three FC layers (64 neurons, 64 neurons, and 8 output neurons) with Tanh activation. The Critic (value network) has three FC layers (64 neurons, 64 neurons, and 1 output neuron) with Leaky ReLU activation. The last FC layer of the Actor model learns a Gaussian distribution. It returns the mean and standard deviation for the actions while the last FC layer of the Critic returns one value that is used to assess the performance of the Actor. Both the shared Conv output and the proprioceptive inputs are normalized before they are fed into the value network and the policy network. This unit utilizes the standard score normalization formula to normalize the input:
$x'=\frac{x-\mu}{\sigma}$,
where $x'$ is the normalized data, $x$ is the original data, $\mu$ is the mean of the input, and $\sigma$ is the standard deviation of the input. This helped with the convergence of the network during training.

At time $t$, reward $r_t$ (more details below) and state $$S_t=4*\{visual\_input, p^{\overrightarrow{j_1}}, p^{\overrightarrow{j_2}}, p^{\overrightarrow{j_3}}, p^{\overrightarrow{g}}\, v^{\overrightarrow{j_1}}, v^{\overrightarrow{j_2}}, v^{\overrightarrow{j_3}}, v^{\overrightarrow{g}}\}$$ (visual and optional proprioceptive/kinesthetic inputs) are generated, then fed into the policy and the value network. Next, the policy network produces the stochastic policy $\pi(A_t|S_t,\theta)$. The actions $A_t$ are sampled from the multidimensional normal distribution. The sampled action vector $A_t$ are fed into the advantage function and used to perform the next action. The value network estimates the value of the state $v^{(S_t,w)}$, which is fed into the advantage function. The advantage function $A^{\pi}(S_t,A_t)$ accumulates the values of $A_t$, $v^{(S_t,w)}$, and the reward $r_t$ for $k$ time steps, then use them to update the policy and value network using the natural gradient for the trust region. The environment receives the sampled action vector $A_t$, takes the next step, and produces the new state and reward ($S_{t+1}$ and $r_{t+1}$).

First, we define subtasks $req_1$ through $req_4$, based on the various body, tool, and target position:
        \nullitem $p^{\overrightarrow{\text{G}}}$ is the pinch position of the gripper,
        \nullitem $p^{\overrightarrow{\text{L}}}$ is the position of the left claw of the gripper,
        \nullitem $p^{\overrightarrow{\text{R}}}$ is the position of the right claw of the gripper,
        \nullitem $p^{\overrightarrow{\text{H}}}$ is the position of the tool handle,
        \nullitem $p^{\overrightarrow{\text{E}}}$ is the position of the tool end-effector,
        \nullitem $p^{\overrightarrow{\text{O}}}$ is the position of the object, and
        \nullitem $p^{\overrightarrow{\text{T}}}$ is the position of the target (arena's bottom).

\begin{itemize}
        \item {\bf Reach-Tool:} The agent's gripper reaches the tool
        \begin{equation}
        req_1=\lVert p^{\overrightarrow{\text{G}}}-p^{\overrightarrow{\text{H}}}  \rVert_2 < \frac{1}{2}gripper\_claw\_length
        \end{equation}
        \item {\bf Grasp-Tool:} The gripper grabs and holds the tool handle
        \begin{eqnarray}
        req_2 =req_1\wedge(close\_theshold&<&\lVert p^{\overrightarrow{\text{L}}}-p^{\overrightarrow{\text{R}}}  \rVert_2 \\ \nonumber
		& < &open\_threshold) 
        \end{eqnarray}
        \item {\bf Tool-Reach-Object:} The tool end-effector reaches the object
        \begin{equation}
        req_3=req_2\wedge(\lVert p^{\overrightarrow{\text{E}}}-p^{\overrightarrow{\text{O}}}  \rVert_2 < \frac{1}{2}tool\_tip\_length)
        \end{equation}
        \item {\bf Object-Reach-Target:} The target object reaches the bottom of the arena
        \begin{equation} \label{eq:req4}
        req_4=\lVert p^{\overrightarrow{\text{O}}}-p^{\overrightarrow{\text{T}}}  \rVert_2 < \frac{1}{2}arena\_boundary\_thickness
        \end{equation}
\end{itemize}

The reward function utilizing the subtasks above are as follows. Note that none of the body, tool, or object positions are provided directly to the agent (these have to be determined by the agent on its own, based on the visual input).

\begin{equation}
r_{\rho}=\begin{cases}
300+k_s*(s_{max}-s_k), & \text{if ($4$)}\\
1.5+1.5*r_{c3}, & \text{if $\neg$($4$)$\wedge$($3$)}\\
0.25+1.25*r_{c2}, & \text{if $\neg$($4$)$\wedge\neg$($3$)$\wedge$($2$)}\\
0.125, & \text{if $\neg$($4$)$\wedge\neg$($3$)$\wedge\neg$($2$)$\wedge$($1$)}\\
0.125*r_{c1}, & \text{otherwise}
\end{cases}
\label{e:r}
\end{equation}
%\end{itemize}
where:
%\begin{itemize}
        \nullitem $\rho=\{p^{\overrightarrow{\text{G}}},p^{\overrightarrow{\text{L}}},p^{\overrightarrow{\text{R}}},p^{\overrightarrow{\text{H}}},p^{\overrightarrow{\text{H}}},p^{\overrightarrow{\text{O}}},p^{\overrightarrow{\text{T}}}\}$,
        \nullitem $k_s$ is a constant used to weight the speed of the task completion. It is set to 3.0 in our experiment,
        \nullitem $s_{max}$ is the maximum time step for each episode (500 steps),
        \nullitem $s_k$ is the time steps so far,
        \nullitem $r_{c1}=1-tanh^2(\frac{\lVert p^{\overrightarrow{\text{G}}}-p^{\overrightarrow{\text{H}}}  \rVert_2}{k_w})$
        \nullitem $r_{c2}=1-tanh^2(\frac{\lVert p^{\overrightarrow{\text{E}}}-p^{\overrightarrow{\text{O}}}  \rVert_2}{k_w})$
        \nullitem $r_{c3}=1-tanh^2(\frac{\lVert p^{\overrightarrow{\text{O}}}-p^{\overrightarrow{\text{T}}}  \rVert_2}{k_w})$
        \nullitem $k_w$ is a constant that is set to the width of the arena,

%One additional change that we made to the tool-use platform is modifying the objective function required to finish the task. Note that there are four requirements to be satisfied to complete the task, including Reach-Tool, Grasp-Tool, and Tool-Reach-Object. There is a problem with requirement 4 (Tool-Reach-Object). Since requirement \ref{eq:req4_orig} includes the requirement \ref{eq:req3_orig} condition, which means that in order to finish the task, the agent has to hold the tool even in the case the target object has already reached the bottom of the arena. We feel this is unnecessary since the tool is merely just a mean to gain access to the target object. Therefore, we removed this condition from requirement \ref{eq:req4_orig} which becomes requirement \ref{eq:req4}. The new objective function is presented as followed:

We trained the networks with the modified ACKTR learning algorithm. ACKTR with Generalized Advantage Estimator (GAE: \citeauthor{schulman2015high} \citeyear{schulman2015high}) computes an estimator of the $k$-step discounted advantage (equation \ref{eq:estimator}) and use equation \ref{eq:gae} to compute the GAE which is the reduced sum of discounted Temporal Difference residuals
\begin{equation} \label{eq:estimator}
\hat{A}_{t}^{(k)}=\sum_{l=0}^{k-1}\gamma^l\delta_{t+l}^V
\end{equation}

\begin{equation} \label{eq:gae}
\hat{A}_{t}^{GAE(\gamma,\lambda)}=\sum_{l=0}^{\infty}(\gamma\lambda)^l\delta_{t+l}^V
\end{equation}
where $\lambda$ is a smoothing parameter used for reducing the variance $0\leq\lambda\leq1$, $\gamma$ is the discount factor constant, and $\delta^{V}_t=r_t+\gamma V(s_{t+1}-V(s_t))$.

The advantage of using Generalized Advantage Estimation is that adjusting the smoothing parameter $\lambda$ might help reduce the variance and lower the biased estimates, which would improve the sample efficiency of Policy Gradient methods significantly.

\section{Experiments and Results}

First, we report the success rate, the mean reward, and the mean episode length of each tool when handled by the trained agent (the same agent was trained to use all three tools). The next step is to quantify the types of behaviors that were displayed by the trained agent while using each tool. The purpose is to examine the environmental condition and the properties of the tools that trigger such behavior. We conducted visual inspection of the episodes to produce a rough interpretation of the agent's strategy in solving the task to support the analysis, and how the tool affordance affected the emergent strategies.

\figtrain

Multiple training sessions were conducted. Each session used twelve workers corresponding to the number of threads. Since the reinforcement learning method that we used is asynchronous, each worker works on a different incarnation of the environment. The number of training threads was divided equally between the types of tools to balance the overall experience provided to the agent. As such, three of the threads were dedicated to training the agent to use the T-shaped tool. Three others were dedicated the L-shaped tool with the end effector pointing to the right and three for the L-shaped tool with the end effector pointing to the left. The rest was for the I-shaped tool.

The number of maximum time steps for an episode was 500, where a time step is equivalent to 10 simulated milliseconds. If the agent finishes its task before reaching the maximum time steps, the remaining time is treated as extra reward. During each training session, the average reward and the average episode length over the threads were recorded. % Low dimensional input, including the positions and velocities of the four joints, are used.

Each training session ran for 19,000 episodes (Intel Core i7-8700K CPU @ 3.7GHz$\times$12, Nvidia GTX 1080-Ti GPU with 11GB memory, RAM 32GB, typical training time of 7 days). An episode is considered to be successful if requirement 4 (equation \ref{eq:req4}) is satisfied, and the total number of time steps is within 500. In all of the five runs, the training successfully converged. On average, training reached a reward of approximately 800 after 19,000 training iterations (note: baseline reward is 300 for just moving the object to the target location; equation \ref{e:r}).

\subsection{Training results}

Figure \ref{fig:perf_five_runs_part1} shows the training results from five training sessions. All of the sessions successfully converged. According to the plot, the training becomes relatively stable after 10,000 episodes with an average reward of approximately 800. To solve the task, the agent must satisfied requirement \ref{eq:req4}. When the agent finishes the task early (less than 500 time-steps), it will receive a bonus that is proportion to the extra time steps. Additionally, figure\ \ref{fig:perf_five_runs_part1} depict the length of the episode of all five runs. Note that these results show the overall performance of the agent using the three given tools, not performance specific to a certain tool. 

\subsection{Quantitative Analysis of Behavior}

\paragraph{Performance by tool type:}
Table\ \ref{t:summary} shows the success rate of the trained agent in handling the three tools over 100 episode runs. An episode is considered as completed if the last requirement (equation \ref{eq:req4}) is satisfied. From the left to the right, the success rate of the T-shaped tool, L-shaped tool, and I-shaped tool are shown. The T-shaped tool has the highest success rate (62\%), followed by the L-shaped tool (50\%), and the I-shaped tool has the lowest success rate (45\%). This result indicates that the T-shaped tool provides the best option to complete the task. This is expected since the T-shaped tool has the shape and physical features that are more advantageous than the other two tools when it comes to dragging the target object from different positions.

Table\ \ref{t:summary} also shows the average episode length of the agent when handling each of the three tools over 100 episode runs. An episode finishes when the agent satisfies the last requirement (equation \ref{eq:req4}) before the time limit is reached (500-time steps). From the left to the right, the episode length when the agent use the T-shaped tool (280.92), the L-shaped tool (302.51), and the I-shaped tool (374.56). Note that the shorter the episode length, the better it is since the agent finishes the task early and get extra reward bonus. Overall, the T-shaped tool enabled the trained agent to solve the task more effectively and faster compared to the other two tools.

\tabsummary

\paragraph{L-shaped Tool (I-like and T-like Use):}
Over the 50 successful episodes of the agent handling the L-shaped tool, we observed that when the tool's end effector points toward the target object the agent exhibit the same hooking behavior similar to the T-shaped tool case (figure\ \ref{f:lmode} Top). However, when the tool's end-effector pointed away from the target object (figure\ \ref{f:lmode} Bottom) the agent usually expresses two interesting behaviors. We named the first behavior as Act-Liked-I behavior to describe the way the agent uses the L-shaped tool in the same manner as the I-shaped tool. It reaches and grasps the tool then uses the body of the tool to wipe down the target object. On the other hand, we named the other behavior as Maneuver behavior (figure\ \ref{f:lcorrect}) where the agent navigate the tool to the other side of the target object so that the tool's end-effector points toward the target object. This way, the agent can effectively use the tool to hook on the target object and drag it down to the bottom of the arena.

Table \ref{t:lshape} presents the quantitative analysis results of the trained agent's behaviors when it handles the L-shaped tool while the tool tip points away from the target object. Over 20 successful episode, 60\% of the time the agent exhibit the Act-Like-I behavior while 40\% of the time, the agent display the Maneuver behavior. Going further into the details, table \ref{t:lshape} shows that when the agent displays the Maneuver behavior, the target object is located in the upper half of the arena 52.5\% of the time and the target object is located in the lower part of the arena 7.5\% of the time. This result indicates that when the target object is located in a position that is high above the agent, the most effective approach to gain access to the target object is to take advantage of the tool's end effector. In this case, the agent will maneuver the tool to hook on to the target object. If the agent uses the other approach (Act-Like-I), the target object might slip out of the control of the tool, thus resulting in a failed episode.

\tablshape

Table \ref{t:lshape} also shows the Act-Like-I behavior versus the position of the target object when the trained agent encounters the L-shaped tool with its end-effector point away from the target object. We can see that 36.67\% of the time the agent display this behavior is when the target object is located in the lower part of the arena and 3.33\% the rest of the time, the target object is located in the upper part of the arena. This can be explained as when the target object position is low, it is easier for the agent to just use the tool to wipe the target object down instead of taking the complicated step of maneuvering the tool to the other side.

\paragraph{I-shaped Tool (Wide Sweep and Hitting):}
Table \ref{t:ishape} shows the break down of the behaviors that occur while the trained agent uses the I-shaped tool. The most common behavior is the Wide-Sweep behavior (happens 71.11\% of the time; figure\ \ref{f:iwide}), followed by the hitting behavior (happens 11.11\% of the time; figure\ \ref{f:ihit}) and other behaviors occur approximately 17.78\% of the time. Since the I-shaped tool is the hardest to use among the three tools (according to the success rate in table \ref{t:summary}) due to the lack of the hook-like tip, the agent has to develop a strategy to compensate for the tool's disadvantages. The wide-sweep behavior is an example in this case as making wide sweep does help with preventing the target object from slipping out of its control. On the other hand, the hitting behavior exhibited by the trained agent while handling the I-shaped tool might indicate the ability to exploit the physics of the environment. This property of the I-shaped tool enables the agent to use the tool tip at the other end to hit the target object down as if using a bat.

Table \ref{t:ishape} also shows these behaviors when using the I-shaped tool in relation to the target object positions. According to the plot, wide-sweep behavior appears the majority of time in the upper part of the arena (55.56\%) and the rest (15.56\%) in the lower part of the arena. When the target object is located high above the agent, it makes sense that the agent has to extend its arm longer to make wider sweep with the target object in order to keep it from slipping. In contrast, the hitting behavior observed in the I-shaped tool scenarios happens most often when the target object is placed in the lower part of the arena (8.89\%) and some times in the upper part (2.22\%). The other less occur behaviors take 17.78\% of the time.

\tabishape

\paragraph{Correcting and Throwing Behaviors:}
Two other less common behaviors that were observed in all types of tools are throwing and correcting. These behavior appear less frequent compared to other major behaviors. The throwing behavior is described as when the agent, after grasping the tool, moves and throws the tool to hit the target object resulting in the target object slide to the bottom of the arena thus completing the task (figure\ \ref{f:ithrow}). This behavior might have been the result of the fact that requirement \ref{eq:req4} in itself does not require that the agent keep holding on to the tool when the target object is already at the bottom. Table \ref{t:throw} shows that the throwing behavior only happens when the trained agent handle the L-shaped and I-shaped tools (2\% for the L-shaped case and 11.11\% for the I-shaped tool case.). The last behavior that is investigated in this study is the Correcting behavior (figure\ \ref{f:lcorrecttwo}). As its name suggests, this behavior indicates that the agent can correct itself nudging the object away and then coming back to drag it. From our experimental results, this behavior happens most frequently when the trained agent uses the T-shaped tool (9.68\% of the time). In the case of the L-shaped tool, it happens 6\% of the time and 11.11\% of the time for the I-shaped tool.

\tabthrow

\subsection{Example Behaviors}

\paragraph{T-shaped Tool (Typical Behavior):}
Figure \ref{f:tdrag} shows a successful episode of the trained agent that uses the T-shaped tool to complete the task. The episode begins when the agent recognizes the tool's location in the environment thanks to the input that it receives. The agent then moves toward the tool and gradually open its grips. In most of the observed cases, to successfully grasp the tool, the agent has to be able to slide its grips to cover the right amount of the tool handle. At this point, it then drives the tool toward the target object. When the agent can use the tool to hook on the target object, it will then drag the target object to the bottom of the arena.

\figtdrag

\paragraph{L-shaped Tool (T-like vs.\ I-like):}
When learning to use the L-shaped tool, the agent exhibits different behaviors in handling the tool. In the case where the tool's end-effector points toward the target object, it displays the same behavior as when it uses the T-shaped tool (see figure \ref{f:lmode} [Top]). However, in the case where the tool's end-effector point away from the target object, it shows two different behaviors. One such behavior is that the agent operates the tool to pull the target object down like the way it uses the I-shaped tool (see figure \ref{f:lmode} [Bottom]). The other behavior is that it maneuvers the tool to the other side of the target object so that the tool's end-effector points toward the target object then hook and pull it down. Figure \ref{f:lcorrect} depicts the behavior.

\figlmode

\figlcorrect

\paragraph{I-shaped Tool (Wide Sweep and Hitting):}
In the case of the I-shaped tool, the agent exhibits two interesting behaviors. One is the wide-sweep where the target object is usually located in the upper part of the arena (figure \ref{f:iwide}; also see figure\ \ref{f:lmode} bottom row, I-like behavior). After grasping the tool ($t=6$), the agent has to extend the tool way beyond the object ($t=10$ to $t=12$) to make a wide swipe ($t=15$ to $t=16$). This ensures that the target object does not slip off the tool during the successive motion. The other behavior is the hitting behavior where the target object is usually located in the lower part of the arena (figure \ref{f:ihit}). In this scenario, after grasping the tool ($t=6$), the agent swings the tool and use its tip to hit the target object ($t=14$), propelling it down.

\figiwide

\figihit

\paragraph{Correcting and Throwing:}
Two behaviors that are not common but can occur sometimes are the correcting behavior and the Throwing behavior. In the case of the correcting behavior, the agent makes a mistake by nudging the object away (figure \ref{f:lcorrecttwo}, $t=10$). However, it circles back and hooks on the object ($t=12$ to $t=14$) and moves it to the other side dragging it to the bottom. In the throwing behavior case
 (figure\ \ref{f:ithrow}), after grasping the tool ($t=9$), the agent moves it for a short distance then throw it toward the target object in an effort to bring it down ($t=15$ to $t=16$). This may happen when the grip on the tool is weak and the agent is about to lose the grip.

\figlcorrecttwo

\figithrow 

\paragraph{Failure Cases:}
Most of the observed failed cases were because the agent could not properly grasp the tool. To grasp the tool correctly, the agent has to approach the tool-handle from a proper direction so that the gripper does not miss and nudge the tool to a distance that the agent cannot reach (figure \ref{f:fail} [Top]). In this case, the correct move is to bend the arm and position the gripper to face the tool handle. Then from this position, the agent should approach the tool to place the tool-handle inside its grippers. Another failed case is when the agent could not control the tool to make the right moves resulting in a wrong way to approach the target object. This usually ends up pushing the target object away from the reach of the agent and its tool. In such circumstances, the result is the agent losing track of its goal, stumbling or freezing until the simulation reaches the limit number of time steps (figure \ref{f:fail} [Middle]). Another unsuccessful scenario is when the agent fails to use the tool properly. For example, in figure \ref{f:fail} (Bottom), when dealing with the I-shaped tool, the agent has to carefully manage the tool so that the target object does not slip out of its hold. In this case, the agent did not extend its arm long enough to make a wider swing, thus failed to move the target object. 

\figfail

\section{Discussion}

The main novelty of our work compared to our prior work \cite{nguyen:ijcnn19} is in the detailed comparison of multiple tool use behavior and quantitative analysis. In our prior work, we focused on adding raw visual input to the agent, simulated physics, and the gripper, with the only tool being the T-shaped tool. Thus, previously we were not able to study the effect of different tool affordances.

Overall, we observed that the agent handles each tool in various manners depending on their affordances. The learned behaviors are specifically aligned with the physical property (shape) of the tool, the environmental condition (the target object position), and also the basic physics of the environment.

In the case of the T-shaped tool, the tool provides the agent with better options to solve the task since it has the hook-like tip facing both ways at the end of the tool end. Therefore, the agent does not have to worry about the object slipping off the tool end. Due to this, the agent can complete the job faster and with higher reward than when using the two other tools.

In the case of the L-shaped tool, the agent exhibited two interesting behaviors when the tool's hook-like tip points away from the target object. This is due to the L-shape having both the T-like (hook) and I-like (no hook) property. It is interesting that the agent could learn to use the tool in both ways, and furthermore, not always strictly dependent on the location of the target object relative to the tool tip direction.

For the I-shaped tool, since it is the most disadvantaged type, more ingenious strategies emerged. These included wide sweep (extending the tool way beyond the object to avoid slipping), hitting (like using a bat), and throwing the tool toward the object. Although the I-shaped tool has lower success rate than the other two, these unique behaviors make it an interesting case study.

%% Figures 

These strategies are quite unique and unexpected, and they are largely dependent on the affordance of the different types of tools, the location of the objects, and also the physics. It is all the more interesting since we used a standard deep RL algorithm, without any direct rewards or motor primitives to for these kinds of behaviors. 

In future work, we intend to analyze the inner workings of our trained neural networks to understand the functional origin of these emergent behaviors.  Are there different modules that emerge in the neural networks, corresponding to the different behaviors? How much do the Conv layers contribute to the assessment of the situation? Also, we will investigate if the learned tool use strategy can be transferable to novel tools (e.g., with a different shape). The tool-body assimilation model in \citet{nishide2011tool} provides a hint. In their work, they trained the model with only the T-shaped and the I-shaped tool, and used the L-shaped tool as a novel tool for testing. We expect our method, when trained and tested in this manner, would show transferability, due to the generalizability of the Conv net.

\section{Conclusion}

In this work, we showed that a single deep reinforcement learning agent can utilize three different types of tools to achieve a common goal, through novel emergent behaviors. The main contribution of this work is to have shown the importance of the type of tool affordance and environmental factors in the emergence of novel behaviors in deep reinforcement learning based control. These behaviors included the use of a single tool in two distinct modes (use of L-shaped as T-shaped or I-shaped), hitting, throwing, correcting, and wide sweep, dependent on tool type and situation. We expect our study to further motivate research on tool use and affordances in deep learning.

%\myack

\bibliography{nguyen}

\section*{Appendix}

\section{Software and Hardware Used}

The following software and versions were used.
\begin{itemize}
	\item MuJuCo v1.31
	\item Python 3.7
	\item OpenAI Gym v0.9.3
	\item OpenAI baselines v0.1.5
\end{itemize}
The hardware used were as follows:
\begin{itemize}
	\item CPU: Intel Core i7-8700K CPU @ 3.7GHz$\times$12
	\item GPU: Nvidia GTX 1080-Ti GPU with 11GB memory, RAM 32GB,
\end{itemize}

\section{Model details}

\begin{table}[h]
\centering
\begin{tabular}{|l|l|l|} \hline 
Layer & Spec & Activation \\ \hline \hline 
Input & 55$\times$48, 3 channels (RGB), 4 frames & \\ \hline
Conv 1&  32 channels @ 8$\times$8 RF, stride = 4 & Leaky ReLU\\ \hline
Conv 2&  64 channels @ 4$\times$4 RF, stride = 2 & Leaky ReLU \\ \hline
Conv 3&  32 channels @ 3$\times$3 RF, stride = 1 & Leaky ReLU\\ \hline
FC & 512 neurons & Leaky ReLU \\ \hline
\end{tabular}
\caption{Shared Convolutional Network. Note: no pooling layer was used.}
\end{table}

\begin{table}[h]
\centering
\begin{tabular}{|l|l|l|} \hline 
Layer & Spec & Activation \\ \hline \hline 
FC layer 1 & 64 neurons & tanh \\ \hline
FC layer 2 & 64 neurons & tanh \\ \hline
FC layer 3 & 8 neurons & tanh \\ \hline
\end{tabular}
\caption{Policy Network.}
\end{table}

\begin{table}[h]
\centering
\begin{tabular}{|l|l|l|} \hline 
Layer & Spec & Activation \\ \hline \hline 
FC layer 1 & 64 neurons & Leaky ReLU\\ \hline
FC layer 2 & 64 neurons & Leaky ReLU\\ \hline
FC layer 3 & 1 neurons & Leaky ReLU\\ \hline
\end{tabular}
\caption{Value Network.}
\end{table}

The parameters for the Kronecker-factored Approximate Curvature (K-FAC) optimizer and Generalized Advantage Estimator (GAE) were as follows:

\begin{itemize}
	\item K-FAC Optimizer: learning rate = 0.25
	\item K-FAC Optimizer: desired KL: 0.001
	\item K-FAC Optimizer: damping factor: 1e-2
	\item GAE discount factor $\gamma$: 0.99
	\item GAE smoothing parameter $\lambda$: 0.95
	\item Number of steps taken (actions sampled) before an update: 30
\end{itemize}

\section{Animations}

These are the animations corresponding to figures 3 to 10 in the main paper. Note: The PDF animation works only in Adobe Acrobat reader.

\animtool{0.8}{7}{T3/T_TL/T_}{3}{30}{T-shaped tool: Typical behavior. Figure 3 in the main paper.}

% Fig 4
\animtool{0.8}{6}{T3/LR_TL/LR_}{3}{30}{L-shaped tool: T-like behavior. Figure 4  (top) in the main paper.}
\animtool{0.8}{6}{T3/LL_TL/LL_}{2}{30}{L-shaped tool: I-like behavior. Figure 4  (bottom) in the main paper.}

% Fig 5
\animtool{0.8}{6}{T3/LLC_TL/LLC_}{2}{30}{L-shaped tool: Changing tool tip direction relative to object. Figure 5 in the main paper.}

% Fig 6
\animtool{0.8}{6}{T3/CR_LL/CR_LL_}{2}{18}{L-shaped tool: Correcting behavior. Figure 6 in the main paper.}

% Fig 7
\animtool{0.8}{6}{T3/I_wide_contact/I_WC_}{2}{16}{I-shaped tool: Wide Sweep behavior. Figure 7 in the main paper.}

% Fig 8
\animtool{0.8}{6}{T3/I_hitting/I_hit_}{2}{16}{I-shaped tool: Hitting behavior. Figure 8 in the main paper.}

% Fig 9
\animtool{0.8}{6}{T3/CR_I/CR_I_}{2}{18}{I-shaped tool: Throwing behavior. Figure 9 in the main paper.}

% Fig 10
\animtool{0.8}{4}{T3/Failed_T/T_}{2}{8}{T-shaped tool: Failure case. Figure 10 (Top) in the main paper.}
\animtool{0.8}{4}{T3/Failed_LL/LL_}{2}{8}{L-shaped tool: Failure case. Figure 10 (Middle) in the main paper.}
\animtool{0.8}{4}{T3/Failed_I/I_}{2}{8}{I-shaped tool: Failure case. Figure 10 (Bottom) in the main paper.}

%\clearpage

\section{Extra Animations}

These are unreported results. 

\animtool{0.8}{2}{T3/FC_I/FC_I_}{3}{6}{I-shaped tool: Failure case. Tool thrown but failed to make object reach the target.}

\animtool{0.8}{3}{T3/FC_T/FC_T_}{2}{6}{T-shaped tool: Failure case. Failed to grab took.}

\animtool{0.8}{5}{T2/c}{2}{30}{Trained with T-shaped tool only. Complex maneuvering and corraling of the object.}

%\animtool{0.8}{3}{T2/f}{2}{12}{Trained with T-shaped tool only. Failure case.}

%\animtool{0.8}{3}{T2/s}{2}{12}{Trained with T-shaped tool only. Failure case.}

\end{document}